\documentclass[11pt,a4paper]{article}
\usepackage[hyperref]{emnlp-ijcnlp-2019}
\usepackage{times}
\usepackage{latexsym}

\usepackage{multirow}
\usepackage{float}
\usepackage{graphicx}
\usepackage{amsmath}
\usepackage{amsfonts}
\usepackage{amssymb}
\usepackage{todonotes}

\usepackage{graphicx}
\usepackage{subcaption}

\usepackage[utf8]{inputenc}
\usepackage[T1]{fontenc}
\usepackage[russian,english]{babel}
\usepackage{booktabs, tabularx}

\usepackage{url}

\aclfinalcopy % Uncomment this line for the final submission

\title{ The Bottom-up Evolution of Representations in the Transformer:\\ A Study with Machine Translation and Language Modeling Objectives}

  \author{Elena Voita$^{1,2}$ \quad Rico Sennrich$^{4,3}$ \quad Ivan Titov$^{3,2}$\bigskip\\
  $^1$Yandex, Russia \quad 
  $^2$University of Amsterdam, Netherlands \\% \smallskip\\
  $^3$University of Edinburgh, Scotland  \quad
  $^4$University of Zurich, Switzerland \\%\smallskip\\
  {\tt lena-voita@yandex-team.ru} \\ {\tt sennrich@cl.uzh.ch} \quad {\tt ititov@inf.ed.ac.uk}
}

\date{}

\begin{document}
\maketitle
\begin{abstract}

We seek to understand how the representations of individual tokens and the structure of the learned feature space evolve between layers in deep neural networks under different learning objectives. We focus on the Transformers for our analysis as they 
 have been shown effective on various tasks, including machine translation (MT),  standard left-to-right language models (LM) and masked language modeling (MLM).  Previous work used black-box probing tasks to show that the representations learned by the Transformer differ significantly depending on the objective. In this work, we use canonical correlation analysis and mutual information estimators to study how information flows across Transformer layers and how this process depends on the choice of learning objective.  
 For example, as you go from bottom to top layers, information about the past in left-to-right language models gets vanished and predictions about the future get formed.  In contrast, for MLM,  representations initially acquire information about the context around the token, partially forgetting the token identity and producing a more generalized token representation. The token identity then gets recreated at the top MLM layers.

\end{abstract}

\section{Introduction}
\label{sect:intro}

Deep (i.e. multi-layered) neural networks have become the standard approach for many natural language processing (NLP) tasks, and their analysis %of neural NLP models 
has been an active topic of research. One popular approach for analyzing representations of neural models is to evaluate how informative they are for various linguistic tasks, so-called ``probing tasks''. Previous work has made some interesting observations regarding these representations; for example, \citet{zhang-bowman-2018-language} show that untrained LSTMs outperform trained ones on a word identity prediction task; and \citet{blevins-etal-2018-deep} show that up to a certain layer performance of representations obtained from a deep LM improves on a constituent labeling task, but then decreases, while with representations obtained from an MT encoder performance continues to improve up to the highest layer. These observations have, however, been somewhat anecdotal and an explanation of the process behind such behavior has been lacking.

In this paper, we attempt to explain more generally why such behavior is observed. 
Rather than measuring the quality of representations obtained from a particular model on some auxiliary task, we characterize how the learning objective determines the information flow in the model. In particular, we 
consider how the representations of individual tokens in the Transformer evolve between layers under different learning objectives.

We look at this task
from the information bottleneck perspective on learning in neural networks. \citet{Tishby2015DeepLA} state that 
``the goal of any supervised learning is to capture and efficiently represent the relevant information in the input variable about the output-label  variable'' and hence the representations undergo transformations which aim to encode as much information about the output label as possible, while `forgetting' irrelevant details about the input. As we study sequence encoders and look into representations of individual tokens rather than the entire input, our situation is more complex. In our model, the information preserved in a representation of a token  is induced due to two roles it plays: (i) predicting the output label from a current token representation;\footnote{We will clarify how we define output labels for LM, MLM and MT objectives in  Section~\ref{sect:task-descriptions}.}
 (ii) preserving information necessary to build representations of other tokens. 
For example, a language model constructs a representation which is not only useful for predicting an output label (in this case, the next token), but also informative for producing representations of subsequent tokens in a sentence.
This is different from the MT setting, 
where there is no single encoder state from which an output label is predicted.
We hypothesize that the training procedure (or, in our notation, the {\bf task}) defines 
\begin{enumerate}
    \item the nature of changes a token representation undergoes, from layer to layer;
    \item the process of interactions and relationships between tokens;
    \item the type of information which gets lost and acquired by a token representation in these interactions.
\end{enumerate}

In this work, we study how the choice of objective affects the process by which information is encoded in token representations of the Transformer~\cite{attention-is-all-you-need}, as this architecture achieves state-of-the-art results on tasks with 
very different 
objectives 
such as machine translation (MT)~\cite{bojar-EtAl:2018:WMT1,iwslt18-overview}, standard left-to-right language modeling (LM)~\cite{radford2018improving} and masked language modeling (MLM)~\cite{bert}.
The Transformer encodes a sentence by iteratively updating representations associated with each token starting from a context-agnostic representation consisting of a positional and a token embedding. At each layer token representations exchange information among themselves via multi-head attention and then this information is propagated to the next layer via a 
feed-forward transformation. We investigate how this process depends on the choice of objective function (LM, MLM or MT) while keeping the data and model architecture fixed.

We start with illustrating the process of information loss and gain in representations of individual tokens by estimating the mutual information between token representations at each layer and the input token identity (i.e.\ the word type) 
or the output label (e.g., the next word for LM).

Then, we investigate behavior of token representations from two perspectives: how they influence and are influenced by other tokens. Using canonical correlation analysis, we  
evaluate the extent of change the representation undergoes and the degree of influence. We reveal differences in the patterns of this behavior for different tasks. 

Finally, we study which type of information gets lost and gained in the interactions between tokens and to what extent a certain property is important for defining a token representation at each layer and for each task. As the properties, we consider token identities (`word type'), positions, identities of surrounding tokens and CCG supertags.
In these analyses we rely on similarity computations.

We find, that
    (1) with the LM objective, as you  go from bottom to top layers, information about the past gets lost and predictions about the future get formed;
    (2) for MLMs, representations initially acquire information about the context around  the  token,  partially  forgetting  the  token identity and producing a more generalized token representation; the token identity then gets recreated at the top layer;
    (3) for MT,  though representations get refined with context, less processing is happening and most information about the word type does not get lost.
This provides us with a hypothesis for why the MLM objective may be preferable in the pretraining context to LM. LMs may not be the best choice, because neither information about current token and its past nor future is represented well: the former since this information gets discarded, the latter since the model does not have access to the future.

Our key contributions are as follows:
\begin{itemize}
    \item we propose to view the evolution of a token representation between layers from the compression/prediction trade-off 
    perspective;

    \item we conduct a series of experiments 
    supporting this view and showing that the two processes, losing information about input and accumulating information about output, take place in the evolution of representations (for MLM, these processes are clearly distinguished and can be viewed as two stages, `context encoding' and `token prediction');
   
    \item we relate to some findings from previous work, putting them in the proposed perspective, and provide insights into the internal workings of Transformer trained with different objectives;
    
    \item we propose an explanation for superior performance of the MLM objective over the LM one for pretraining.

\end{itemize}

All analysis is done in a model-agnostic manner by investigating properties of token representations at each layer, and can, in principle, be applied to other multi-layer deep models (e.g., multi-layer RNN encoders).

\section{Tasks}
\label{sect:tasks}
\label{sect:task-descriptions}
\vspace{-1ex}

In this section, we describe the 
tasks we consider. 
For each task, we define  input $X$ and output $Y$.

\subsection{Machine translation}
\label{sect:tasks_mt}

Given source sentence $X\!=\!(x_1, x_2, ..., x_S)$ and a target sentence $Y\!=\!(y_1, y_2, ..., y_T)$, NMT models predict words in the target sentence, word by word, i.e. provide estimates of the conditional distribution $p(y_i| X, y_{1, i-1}, \theta)$.

We train a standard 
Transformer for the translation task and then analyze its encoder. 
In contrast to the other two tasks we describe below, representations from the top layers are not directly used to predict output labels but to
encode the information which is then used by the decoder.

\subsection{Language modeling}
\label{sect:tasks_clm}

LMs estimate the probability of a word given the previous words in a sentence~$P(x_t|x_1, \dots, x_{t-1}, \theta)$. More formally, the model is trained with inputs $X=(x_1, \dots, x_{t-1})$ and outputs $Y=(x_t)$, where $x_t$ is the output label predicted from the final (i.e.\ top-layer) representation of a token $x_{t-1}$.
It is straightforward to apply the Transformer to this task \citep{radford2018improving,lample&conneau2019_pretraining}.

\subsection{Masked language modeling}
\label{sect:tasks_mlm}
%\vspace{-1ex}

We also consider the MLM objective~\cite{bert}, randomly sampling $15\%$ of the tokens to be predicted. 
 We replace the corresponding input token by [MASK] or a random token in $80\%$ and $10\%$ of the time, respectively, keeping it unchanged otherwise. 

For a sentence $(x_1, x_2, ..., x_S)$, where token $x_i$ is replaced with  $\tilde{x_i}$, the model receives  $X\!=\!(x_1, \dots, x_{i-1}, \tilde{x_i}, x_{i+1}, \dots, x_S)$ as input and needs to predict $Y\!=\!(x_i)$. The label $x_i$ is predicted from the final representation of the token~$\tilde{x_i}$. % after encoding.

\section{Data and Setting}
\label{sect:data-and-setting}

As described below, for a fair comparison,  we use the same training data, model architecture and parameter initialization across all the tasks.  
In order to make sure that our findings are reliable,
we also use multiple datasets and repeat experiments with different random initializations for each task.

We train all models on the data from the WMT news translation shared task. We conduct separate series of experiments using two language pairs: WMT 2017 English-German  (5{.}8m 
sentence pairs) and WMT 2014 English-French (40{.}8m
sentence pairs). For language modeling, we use only the source side of the
parallel data.
We remove randomly chosen 2{.}8m sentence pairs from the English-French dataset and use the source side for analysis. English-French models are trained on the remaining 38m sentence pairs. We consider different dataset sizes (2{.}5m and 5{.}8m for English-German, 2{.}5m, 5{.}8m and 38m for English-French). We find that our findings are true for all languages, dataset sizes and initializations. In the following, all the illustrations are provided for the models trained on the full English-German dataset (5{.}8m sentence pairs).

We follow the setup and training procedure of the Transformer base model~\cite{attention-is-all-you-need}. For details, see the appendix.

\section{The Information-Bottleneck Viewpoint}
\label{sect:mutual_info}
\vspace{-1ex}

In this section, we give an intuitive explanation of the Information Bottleneck (IB) principle~\cite{tishby2000information}  and consider a direct application of this principle to our analysis.

\subsection{Background}
\label{sect:ib_formal}

The IB method~\cite{tishby2000information} considers a joint distribution of input-output pairs $p(X, Y)$ and aims to extract a compressed representation
$\tilde{X}$ for an input $X$ such that $\tilde{X}$ retains as much as possible information  about the output~$Y$. More formally, the IB method maximizes the mutual information (MI) with the output  $I(\tilde{X};Y)$, while penalizing for  MI with the input $I(\tilde{X};X)$. The latter term in the objective ensures that the representation is indeed a compression. Intuitively, the choice of the output variable $Y$ determines the split of $X$ into irrelevant and relevant features. The relevant features need to be retained while irrelevant ones should be dropped. 

\citet{Tishby2015DeepLA} argue that computation in a multi-layered neural model can be regarded as an evolution towards the theoretical optimum of the IB objective. 
A sequence of layers is viewed as a Markov chain, and the  process of obtaining $Y$ corresponds to 
compressing the representation
as it flows across layers and retaining only information relevant to predicting $Y$. This implies that $Y$ defines the information flow in the model. 
Since $Y$ is different for each model, we expect to see different patterns of information flow in models, and this is the  focus of our study.

\subsection{IB for token representations}
\label{sect:ib_relation_to_our_work}

In this work, we view every sequence model (MT, LM and MLM) as learning a function from input $X$ to output $Y$. The input is a sequence of tokens $X=(x_1, x_2, \dots, x_n)$ and the output $Y$ is defined in Section~\ref{sect:tasks}.
Recall that we focus on representations of individual tokens in every layer rather than the representation of the entire sequence.

We start off our analysis of divergences in the information flow for different objectives by estimating the amount of information about input or output tokens retained in the token representation at each layer.

\subsubsection{Estimating mutual information}

Inspired by \citet{Tishby2015DeepLA}, 
we estimate MI between token representations at a certain layer and an input token. To estimate MI, we need to consider token representations at a layer as samples from a discrete distribution. To get such distribution, in the original works~\cite{ib_visualization}, the authors binned the neuron's $\arctan$ activations. Using these discretized values for each neuron in a layer, they were able to treat a layer representation as a discrete variable.
They considered neural networks with maximum 12 neurons at a layer, but in practical scenarios (e.g.\ we have 512 neurons in each layer) this approach is not feasible. Instead, similarly to~\citet{NIPS2018_7769}, we discretize the representations by clustering them into a large number of clusters. Then we use cluster labels instead of the continuous representations in the MI estimator. 

Specifically, we take only representations of the 1000 most frequent (sub)words. We gather representations for 5 million occurrences of these at each layer for each of the three models. We then cluster the representations into $N=10000$ clusters using mini-batch
$k$-means with $k = 100$. In the experiments studying the mutual information between a layer and source (or target) labels we further filter occurrences.
Namely, we take only occurrences where the source and target labels are among the top 1000 most frequent (sub)words.

\subsubsection{Results}

\begin{figure}[t!]
\center{\includegraphics[scale=0.25]{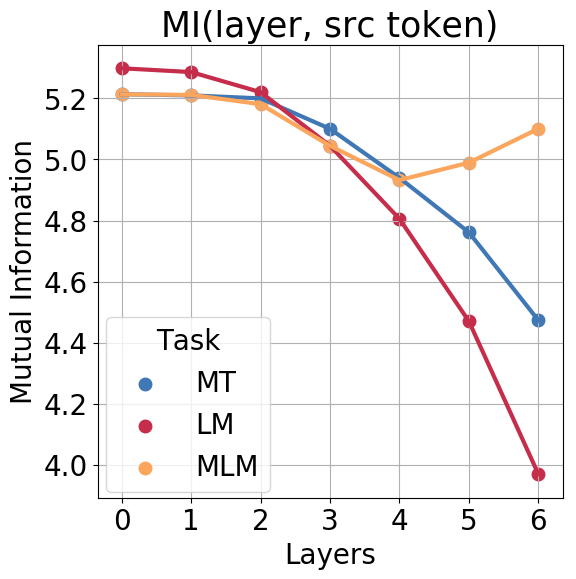}}
\vspace{-1ex}
\caption{The mutual information between an input token and a representation of this token at each layer.}
\vspace{-2ex}
\label{fig:mutual_info}
\end{figure}

\begin{figure}[t!]
    \centering

    \begin{subfigure}[b]{0.19\textwidth}
        \includegraphics[width=\textwidth]{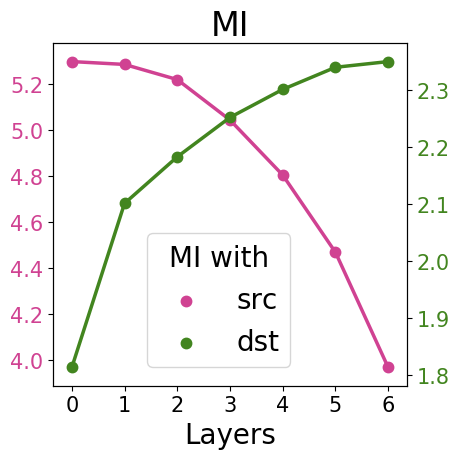}
        \vspace{-3ex}
        \caption{LM}
    \end{subfigure}
    \quad
    \begin{subfigure}[b]{0.19\textwidth}
        \includegraphics[width=\textwidth]{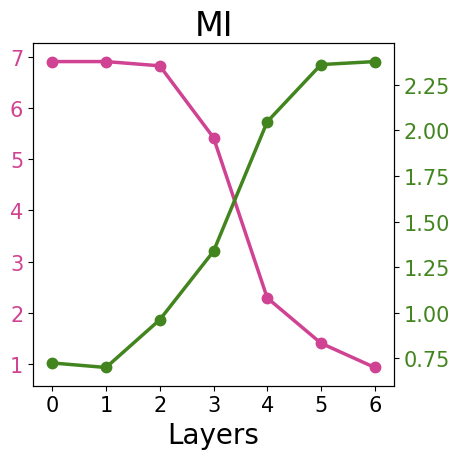}
        \vspace{-3ex}
        \caption{MLM}
    \end{subfigure}
    \vspace{-2ex}
    \caption{The mutual information of token representations at a layer and source (or target) tokens. For MLM, only tokens replaced at random are considered to get examples where input and output are different.}
    \vspace{-2ex}
    \label{fig:ib_mi_src_dst}
\end{figure}

First, we estimate the MI between an input token and a representation of this token at each layer. In this experiment, we form data for MLM as in the test regime; in other words, the input token is always the same as the output token. Results are shown in Figure~\ref{fig:mutual_info}.
For LM, the amount of relevant information about the current input token decreases. This agrees with our expectations:
some of the information about the history is intuitively not  relevant for predicting the future. 
MT shows a similar behavior, but the decrease is much less sharp.
This is again intuitive: the information about the exact identity is likely to be useful for the decoder.  
The most interesting and surprising graph is for MLM: first, similarly to other models, the information about the input token is getting lost but then, at two top layers, it gets recovered. We will refer to these phases in further discussion as context encoding and token reconstruction, respectively. 
Whereas such non-monotonic behavior is impossible when analyzing entire layers, as in \citet{Tishby2015DeepLA}, in our case,  it suggests that this extra information is obtained from other tokens in the context. 

We perform the same analysis but now measuring MI with the output label for LM and MLM. In this experiment, we form data for MLM as in training, masking or replacing a fraction of tokens. We then take only tokens replaced with a random one to get examples where input and output tokens are different. Results are shown in Figure~\ref{fig:ib_mi_src_dst}. 
We can see that, as expected, MI with input tokens decreases while MI with output tokens increases. Both LM and MLM are trained to predict a token (next for LM and current for MLM) by encoding input and context information. 
While in Figure~\ref{fig:mutual_info} we observed monotonic behavior of LM, when looking at the information with both input and output tokens, we can 
see the two processes, losing information about input and accumulating information about output, for both LM and MLM models. For MLM these processes are more distinct and can be thought of as
the context encoding and token prediction (compression/prediction) stages. 
For MT, since nothing is predicted directly, we see only the encoding stage of this process. This observation relates also to the findings by~\citet{blevins-etal-2018-deep}. They show that up to a certain layer the performance of representations obtained from a deep multi-layer RNN LM improves on a constituent labeling task, but then decreases, while for representations obtained from an MT encoder performance continues to improve up to the highest layer. We further support this view with other experiments in Section~\ref{sect:evolution_past_future}.

Even though the information-theoretic view provides insights into processes shaping the representations, direct MI estimation from finite samples for densities on multi-dimensional spaces   is challenging~\cite{paninski2003estimation}.
For this reason in the subsequent analysis we 
use more well-established frameworks such as  canonical correlation analysis to provide new insights and also to corroborate  findings we made in this section (e.g., the presence of two phases in MLM encoding). Even though we will be using different machinery,  we will focus on the
same two IB-inspired questions: (1) how does information flow across layers? and (2) what information does a layer represent?

\section{Analyzing Changes and Influences}
\label{sect:pwcca_intro}
\vspace{-1ex}

In this section, we 
analyze the flow of information.
The questions we 
ask 
include: how much processing is happening in a given layer; which tokens influence other tokens most; which tokens gain most information from other tokens.
As we will see, these questions can be reduced to a comparison between network representations.
We start by describing the tool we use.

\subsection{Canonical Correlation Analysis}

We rely on recently introduced projection weighted Canonical Correlation Analysis (PWCCA)~\cite{pwcca-2018}, which is an improved version of SVCCA~\cite{svcca-2017}. Both approaches are based on classic Canonical Correlation Analysis~(CCA) \cite{cca}.

CCA is a multivariate statistical method for relating two sets of observations arising from an underlying process. In our setting, the underlying process is a neural network trained on some task. The two sets of observations
can be seen as `two views' on the data. 
Intuitively, we look at the same data (tokens in a sentence) from two standpoints. For example, one view is one layer and another view is another layer. 
Alternatively, one view can be $l$-th layer in one model, whereas another view can be the same $l$-th layer in another model. CCA lets us measure similarity between pairs of 
views. 

Formally, given a set of tokens $(x_1, x_2, \dots, x_N)$ (with the sentences they occur in), we gather their representations produced by two models ($m_1$ and $m_2$) at layers $l_1$ and $l_2$, respectively. To achieve this, we encode the whole sentences and take representations of  tokens we are interested in. We  get two views of these tokens by the models: 
$v^{m_1,l_1}\!=\!(x_1^{m_1, l_1}, x_2^{m_1, l_1}, \dots, x_N^{m_1, l_1})$ and $v^{m_2,l_2}\!=\!(x_1^{m_2, l_2}, x_2^{m_2, l_2}, \dots, x_N^{m_2, l_2})$. The representations are gathered in two matrices $X_1\!\in\! M_{a, N}$ and $X_2\!\in\! M_{b, N}$, where $a$ and $b$ are the numbers of neurons in the models.\footnote{In our experiments, $N>100000$.} These matrices are then given to CCA (specifically, PWCCA). CCA identifies a  linear relation maximizing the correlation between the matrices and computes the similarity. The values of PWCCA range from 0 to 1, with 1 indicating that the observations are linear transformations of each other, 0 indicating no correlation.

In the next sections, we vary two aspects of this process: tokens 
and 
the `points of view'.

\subsection{A coarse-grained view}
We start with the analysis where we do not attempt to distinguish between different token types.

\subsubsection{Distances between tasks}

\begin{figure}[t!]
    \centering
    \begin{subfigure}[b]{0.27\textwidth}
    \vspace{-1ex}
        \includegraphics[width=\textwidth]{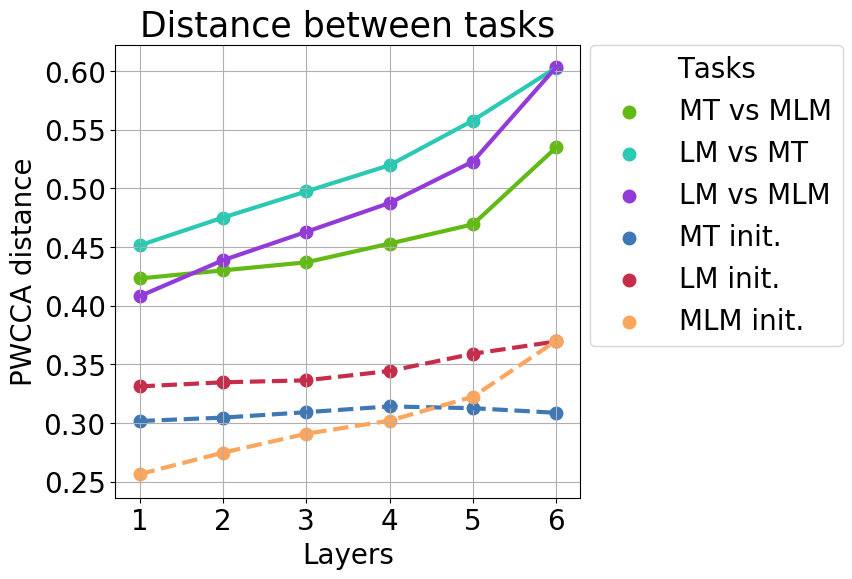}
        \caption{}
        \label{fig:general_between_tasks}
    \end{subfigure}
    \begin{subfigure}[b]{0.18\textwidth}
    \vspace{-1ex}
        \includegraphics[width=\textwidth]{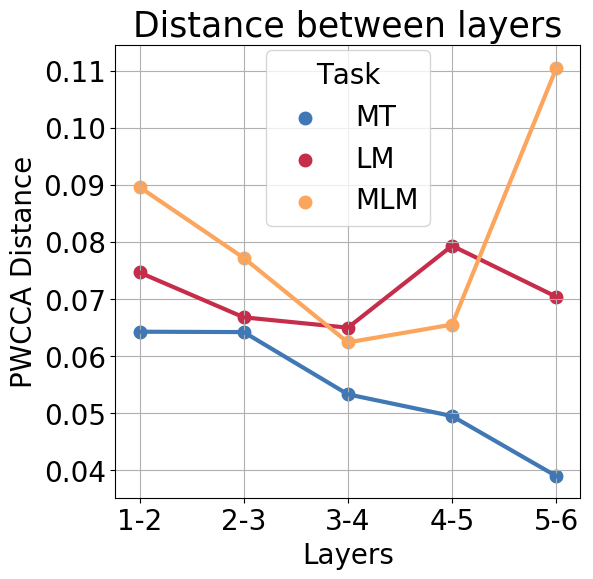}
        \caption{}
        \label{fig:general_between_layers}
    \end{subfigure}
    \vspace{-1ex}
    \caption{PWCCA distance (a) between representations of different models at each layer (``init.'' indicates different initializations), (b) between consecutive layers of the same model.}
    \vspace{-2ex}
\end{figure}

As the first step in our analysis, we measure the difference between representations learned for different tasks. In other words, we compare representations for $v^{m_1,l}$ and $v^{m_2,l}$ at different layers $l$.
Here the data is all tokens of 5000 sentences.
We also quantify differences between representations of models trained with the same objective  but different random initializations. The results are provided in Figure~\ref{fig:general_between_tasks}. 

First, differences due to training objective are much larger than the ones due to random initialization of a model. This indicates that PWCCA captures underlying differences in the types of information learned by a model rather than those due to randomness in the training process. 

MT and MLM objectives produce representations that are closer to each other than to LM's representations. The reason for this might be two-fold. 
First, for LM only preceding tokens are in the context, whereas for MT 
and MLM it is the entire sentence.
Second, both MT and MLM focus on a given token, as it either needs to be reconstructed or translated. In contrast, LM produces a representation needed for predicting the next token.

\subsubsection{Changes between layers}

In a similar manner, we measure the difference between representations of consecutive layers in each model (Figure~\ref{fig:general_between_layers}). In this case we take views $v^{m,l}$ and $v^{m,l+1}$ and vary layers $l$ and tasks $m$.

For MT,  the extent of change monotonically decreases when going from the bottom to top layers, whereas there is no such monotonicity for LM and MLM.  This mirrors our view of LM and especially MLM as undergoing phases of encoding and reconstruction (see Section~\ref{sect:ib_relation_to_our_work}), 
thus requiring a stage of dismissing information irrelevant to the output, which, in turn, is accompanied by large changes in the representations between 
layers.

\subsection{Fine-grained analysis}

In this section, we select tokens with some predefined property (e.g., frequency) and investigate how much the tokens are influenced by other tokens or 
how much they influence other tokens.

\noindent
{\bf Amount of change.} We measure the extent of change for a group of tokens as the PWCCA distance between the representations of \underline{these tokens} for a pair of adjacent layers $(l, l+1)$. This quantifies the amount of information the tokens receive in this layer.

\noindent
{\bf Influence.}
To measure the influence of a token at $l$th layer on other tokens, we measure PWCCA distance between two versions of representations of \underline{other tokens} in a sentence: first after encoding as usual, second when encoding first $l-1$ layers as usual and masking out the influencing token 
at the $l$th layer.\footnote{By masking out we mean that other tokens are forbidden to attend to the chosen one.}

\subsubsection{Varying token frequency}
\label{sec:pwcca_token_frequency}

\begin{figure}[t!]
    \centering
    \begin{subfigure}[b]{0.23\textwidth}
        \includegraphics[width=\textwidth]{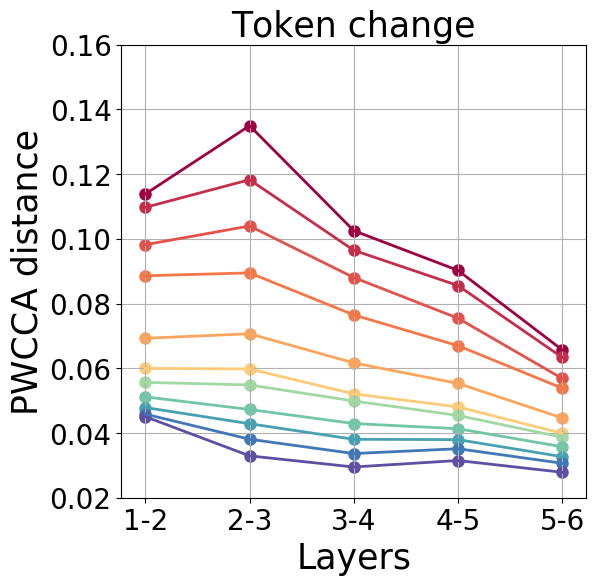}
        \caption{MT}
        \label{fig:change_by_frequency_mt}
    \end{subfigure}
    \begin{subfigure}[b]{0.23\textwidth}
        \includegraphics[width=\textwidth]{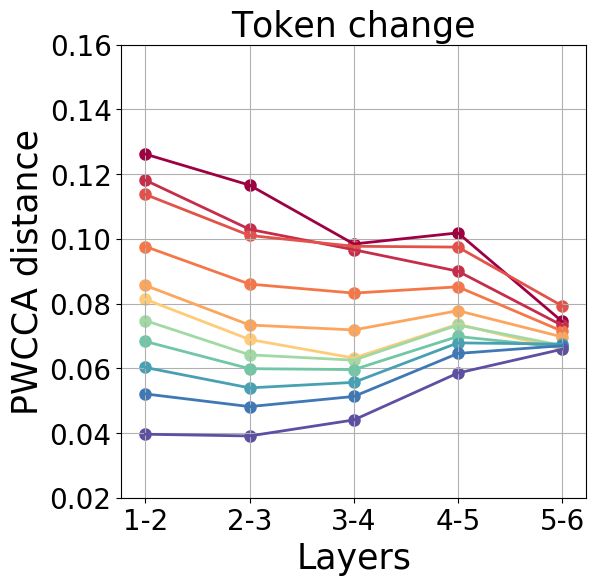}
        \caption{LM}
        \label{fig:change_by_frequency_lm}
    \end{subfigure}
    \begin{subfigure}[b]{0.23\textwidth}
        \includegraphics[width=\textwidth]{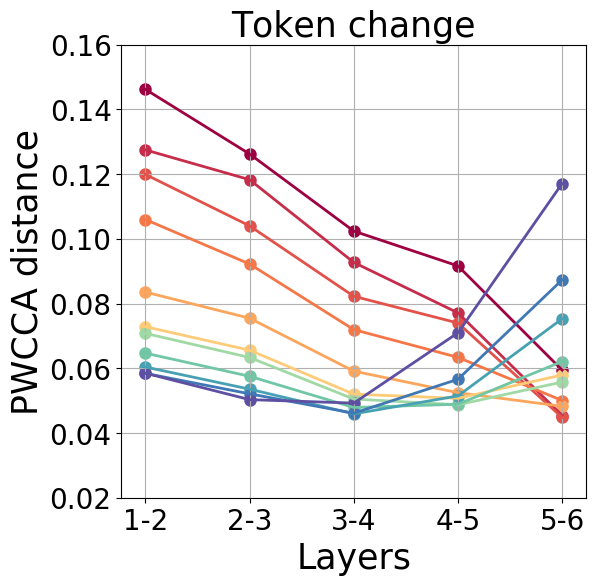}
        \caption{MLM}
        \label{fig:change_by_frequency_mlm}
    \end{subfigure}
    \quad
    \begin{subfigure}[b]{0.17\textwidth}
        \includegraphics[width=\textwidth]{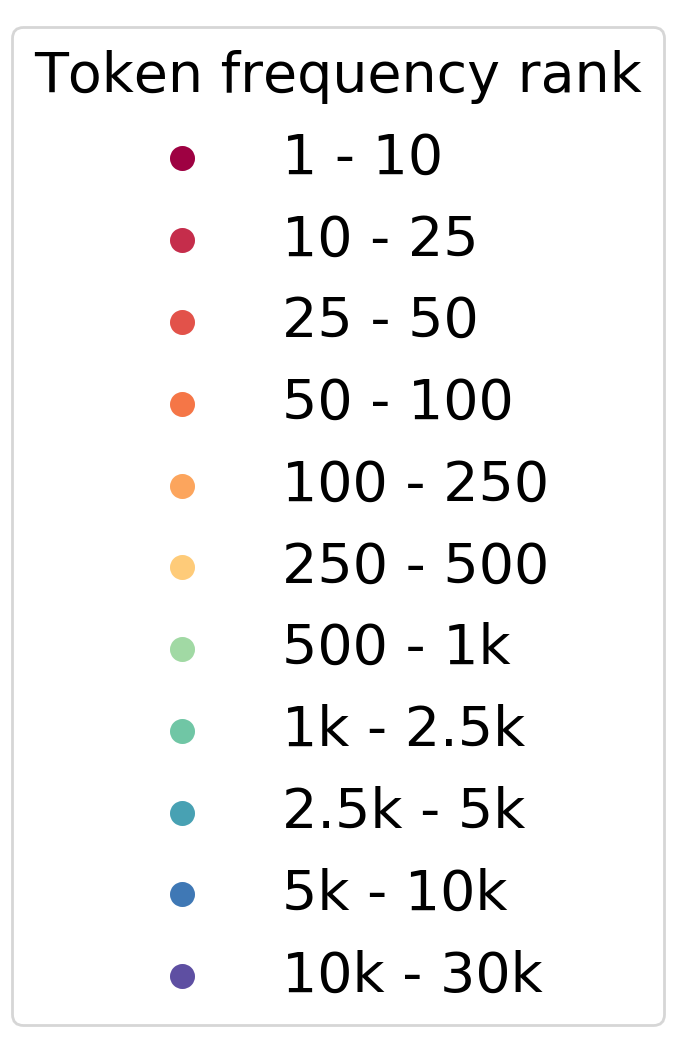}
    \end{subfigure}
    \vspace{-2ex}
    \caption{Token change vs its frequency.}
    \vspace{-2ex}
    \label{fig:change_by_frequency}
\end{figure}

Figure~\ref{fig:change_by_frequency} shows a clear dependence
of the amount of change on token frequency. Frequent tokens change more than rare ones in all layers in both LM and MT.  Interestingly, unlike MT, for LM this dependence 
dissipates as we move towards top layers. We can speculate that top layers focus on predicting the future rather than incorporating the past, and, at that stage, token frequency of the last observed token becomes less important.

The behavior for MLM is quite 
different. The  
two stages for MLMs  could already be seen in Figures \ref{fig:mutual_info}  and \ref{fig:general_between_layers}. They are even more pronounced here.  
The transition from a generalized token representation, formed at the encoding stage, to recreating token identity apparently requires more changes for rare tokens. 

\begin{figure}[t!]
    \centering
    \begin{subfigure}[b]{0.23\textwidth}
        \includegraphics[width=\textwidth]{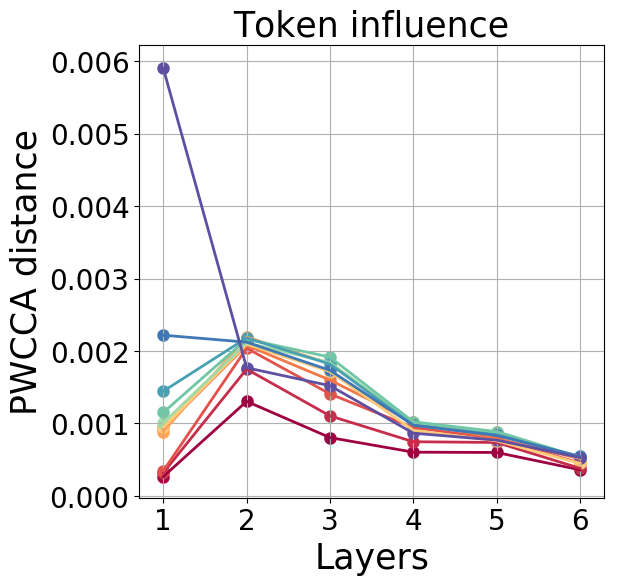}
        \caption{MT}
        \label{fig:change_by_frequency_mt}
    \end{subfigure}
    \begin{subfigure}[b]{0.23\textwidth}
        \includegraphics[width=\textwidth]{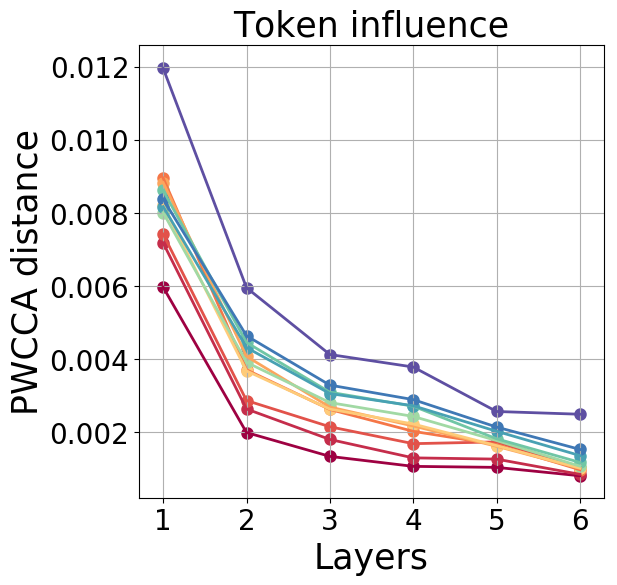}
        \caption{LM}
        \label{fig:change_by_frequency_lm}
    \end{subfigure}
    \begin{subfigure}[b]{0.23\textwidth}
        \includegraphics[width=\textwidth]{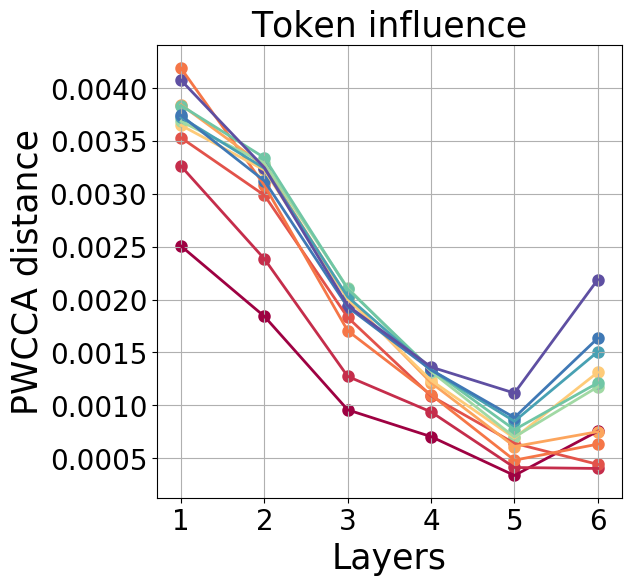}
        \caption{MLM}
        \label{fig:change_by_frequency_mlm}
    \end{subfigure}
    \quad
    \begin{subfigure}[b]{0.17\textwidth}
        \includegraphics[width=\textwidth]{pict/freq_wmt_en_de_bbox.png}
    \end{subfigure}
    \vspace{-1ex}
    \caption{Token influence vs its frequency.}
    \vspace{-2ex}
    \label{fig:influence_by_frequency}
\end{figure}

\begin{figure}[t!]
    \centering
    \begin{subfigure}[b]{0.23\textwidth}
        \includegraphics[width=\textwidth]{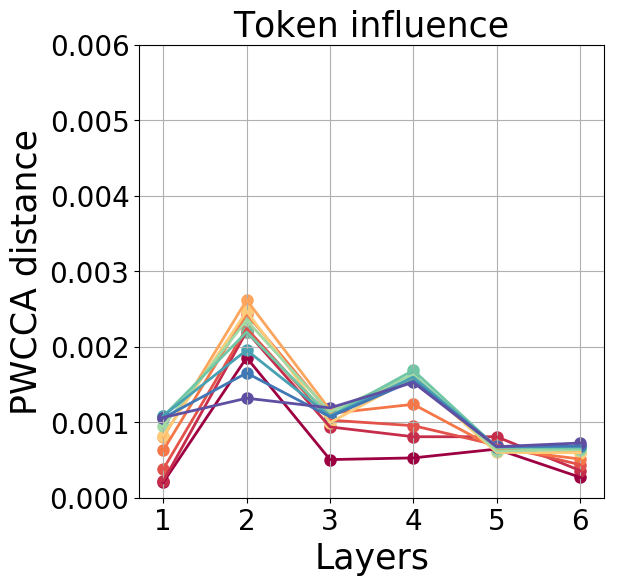}
        \caption{MT}
        \label{fig:change_by_frequency_mt_iwd}
    \end{subfigure}
    \begin{subfigure}[b]{0.23\textwidth}
        \includegraphics[width=\textwidth]{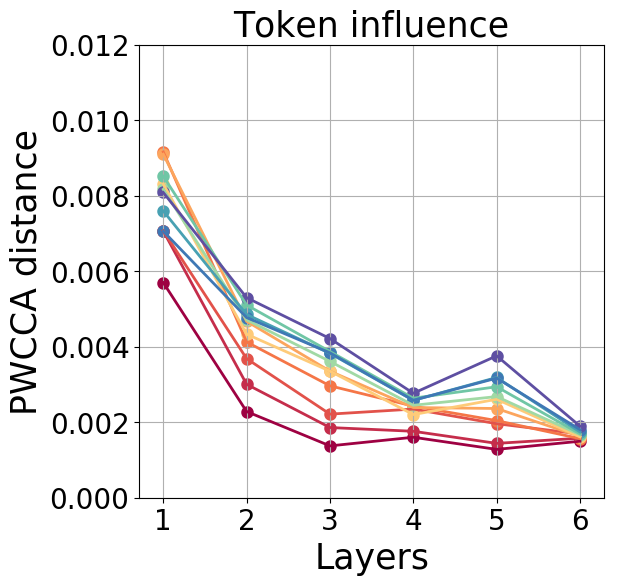}
        \caption{LM}
        \label{fig:change_by_frequency_lm_iwd}
    \end{subfigure}
    \vspace{-1ex}
    \caption{Token influence vs its frequency for models trained with word dropout (in training, each input token is replaced with a random with the probability 10$\%$).}
    \vspace{-2ex}
    \label{fig:influence_by_frequency_iwd}
\end{figure}

When measuring influence, we find that rare tokens generally influence more than frequent ones (Figure~\ref{fig:influence_by_frequency}). We notice an extreme influence of rare tokens at the first MT layer and at all LM layers. In contrast, rare tokens are not the most influencing ones at the lower layers of MLM. 
We hypothesize that the training procedure of MLM, with masking out some tokens or replacing them with random ones, teaches the model not to over-rely on these tokens before their context is well understood. To test our hypothesis, we additionally trained MT and LM models with token dropout on the input side (Figure~\ref{fig:influence_by_frequency_iwd}).  
As we expected, there is no extreme influence of rare tokens when using this regularization, supporting the above interpretation. 

Interestingly, 
our earlier study of the MT Transformer \cite{voita-etal-2019-analyzing} shows how
this influence of rare tokens is implemented by the model.
In that work, we observed that, for any considered language pair, there is one dedicated attention head in the first encoder layer which tends to point to the least frequent tokens in every sentence. 
The above analysis suggest that this phenomenon is likely due to  overfitting.

We also analyzed the extent of change and influence splitting tokens according to their part of speech; see appendix for details.

\section{What does a layer represent?}

Whereas in the previous section we were interested in quantifying the amount of information exchanged between tokens, here we primarily want to understand what representation in each layer `focuses' on. 
We evaluate to what extent a certain property is important for defining a token representation at each layer by (1) selecting a large number of token occurrences and taking their representations; (2) validating if a value of the property is the same for token occurrences corresponding
to the closest representations.
Though our approach is different from probing tasks, we choose the properties which will enable us to relate to other works reporting similar behaviour~\cite{zhang-bowman-2018-language,blevins-etal-2018-deep,tenney-etal-2019-bert}. 
The properties we consider are token identity, position in a sentence, neighboring tokens and CCG supertags.

\subsection{Methodology}

For our analysis, we take 
100 random word types from the top 5,000 in our vocabulary. For each word type, 
we gather 1,000 different occurrences along with the representations from all three models. For each representation, we take the closest neighbors among representations at each layer and evaluate the percentage of neighbors with the same value of the property.

\subsection{Preserving token identity and position}
\label{sec:evolution_same_token}

In this section, we track the loss of information about token identity (i.e., word type) and position. Our motivation is three-fold. First, this will help us to confirm the results provided on Figure~\ref{fig:mutual_info}; second, to relate to the work reporting results for probing tasks predicting token identity. Finally, the Transformer starts encoding a sentence from a positional and a word embedding, thus it is natural to look at how this information is preserved. 

\begin{figure}[t!]
    \centering
    \begin{subfigure}[b]{0.20\textwidth}
        \includegraphics[width=\textwidth]{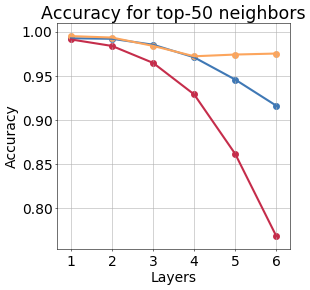}
        \caption{}
        \label{fig:same_token}
    \end{subfigure}
    \quad
    \begin{subfigure}[b]{0.23\textwidth}
         \includegraphics[width=\textwidth]{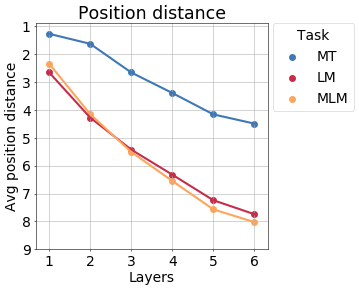}
        \caption{}
        \label{fig:same_position}
    \end{subfigure}
    \vspace{-1ex}
    \caption{Preserving (a) token identity, (b) position}
    %\vspace{-1ex}
\end{figure}

\begin{figure}
\begin{tabular}{ccccccc} %number of "c"s is row length
 \!\!\includegraphics[width=8mm]{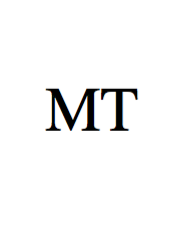}\!\!\! & \!\!\includegraphics[width=11.5mm]{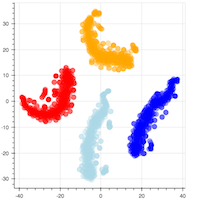}\!\!\! & \!\!\!\!\includegraphics[width=11.5mm]{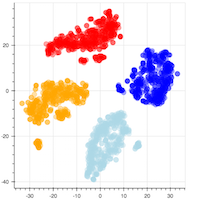}\!\!\! & \!\!\!\!\includegraphics[width=11.5mm]{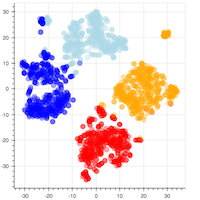}\!\!\! & \!\!\!\!\includegraphics[width=11.5mm]{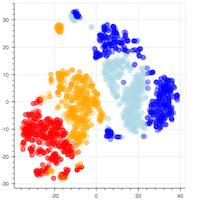}\!\!\! & 
    \!\!\!\!\includegraphics[width=11.5mm]{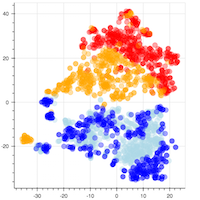}\!\!\! &
    \!\!\!\!\includegraphics[width=11.5mm]{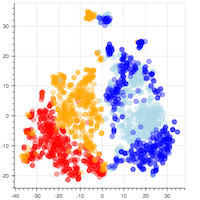} \\
    
 \!\!\includegraphics[width=8mm]{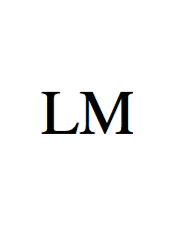}\!\!\! & \!\!\includegraphics[width=11.5mm]{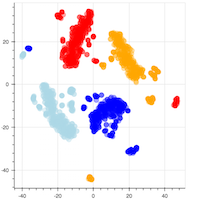}\!\!\! & \!\!\!\!\includegraphics[width=11.5mm]{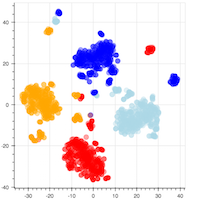}\!\!\! & \!\!\!\!\includegraphics[width=11.5mm]{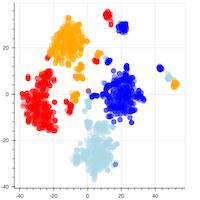}\!\!\! & \!\!\!\!\includegraphics[width=11.5mm]{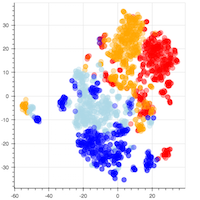}\!\!\! & 
    \!\!\!\!\includegraphics[width=11.5mm]{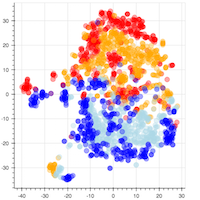}\!\!\! &
    \!\!\!\!\includegraphics[width=11.5mm]{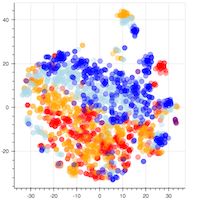} \\
    
 \!\!\includegraphics[width=8mm]{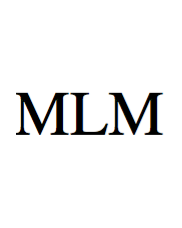}\!\!\! & \!\!\includegraphics[width=11.5mm]{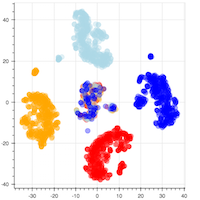}\!\!\! & \!\!\!\!\includegraphics[width=11.5mm]{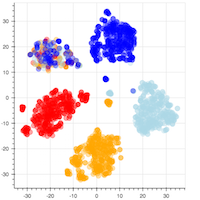}\!\!\! & \!\!\!\!\includegraphics[width=11.5mm]{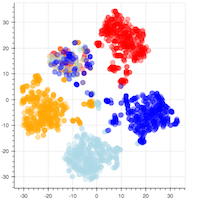}\!\!\! & \!\!\!\!\includegraphics[width=11.5mm]{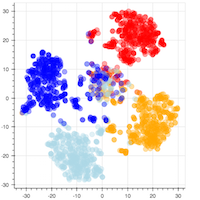}\!\!\! & 
    \!\!\!\!\includegraphics[width=11.5mm]{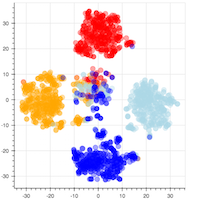}\!\!\! &
    \!\!\!\!\includegraphics[width=11.5mm]{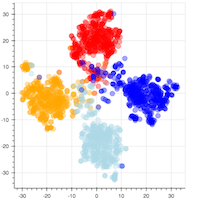} \\
\end{tabular}
\vspace{-1ex}
\caption{t-SNE of different occurrences of the tokens ``is'' (red), ``are'' (orange), ``was'' (blue), ``were'' (lightblue). On the x-axis are layers.}
\vspace{-1ex}
\label{fig:tsne_be_tokens}
\end{figure}

\subsubsection{Preserving token identity}

We want to check to what extent a model confuses representations of different words. For each of the considered words we add 9000 representations of words which potentially could be confused with it.\footnote{See appendix for the details.} For this extended set of representations, we follow the methodology described above.

Results are presented in Figure~\ref{fig:same_token}. Reassuringly, the plot is very similar to the one computed with MI estimators (Figure~\ref{fig:mutual_info}), further supporting the interpretations we gave previously (Section~\ref{sect:mutual_info}). Now, let us recall the findings by~\citet{zhang-bowman-2018-language} regarding the superior performance of untrained LSTMs over trained ones on the task of token identity prediction. They mirror our view of the evolution of a token representation as going through compression and prediction stages, where the learning objective defines the process of forgetting information. If a network is not trained, it is not forced to forget input information.

Figure~\ref{fig:tsne_be_tokens} shows how representations of different occurrences of the words ``is'', ``are'', ``was'', ``were'' get mixed in MT and LM layers and disambiguated for MLM. For MLM, $15\%$ of tokens were masked as in training. In the first layer, these masked states form a cluster separate from the others, and then they get disambiguated as we move bottom-up across the layers.

\subsubsection{Preserving token position}

\begin{figure}
\begin{tabular}{ccccccc} 
 \!\!\includegraphics[width=8mm]{tsne/mt.png}\!\!\! & \!\!\includegraphics[width=11.5mm]{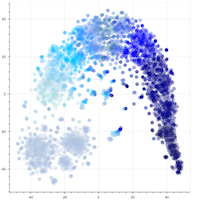}\!\!\! & \!\!\!\!\includegraphics[width=11.5mm]{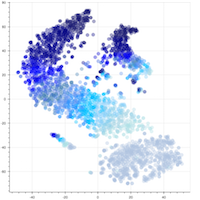}\!\!\! & \!\!\!\!\includegraphics[width=11.5mm]{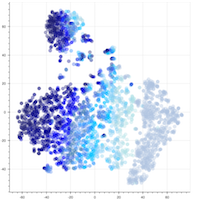}\!\!\! & \!\!\!\!\includegraphics[width=11.5mm]{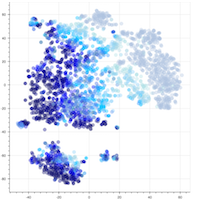}\!\!\! & 
    \!\!\!\!\includegraphics[width=11.5mm]{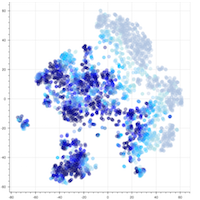}\!\!\! &
    \!\!\!\!\includegraphics[width=11.5mm]{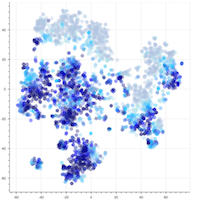} \\
    
 \!\!\includegraphics[width=8mm]{tsne/lm.png}\!\!\! & \!\!\includegraphics[width=11.5mm]{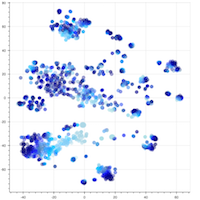}\!\!\! & \!\!\!\!\includegraphics[width=11.5mm]{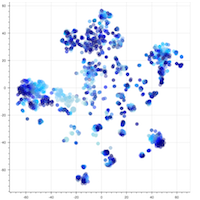}\!\!\! & \!\!\!\!\includegraphics[width=11.5mm]{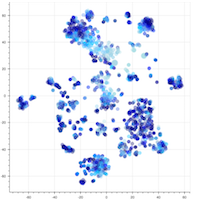}\!\!\! & \!\!\!\!\includegraphics[width=11.5mm]{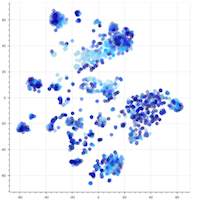}\!\!\! & 
    \!\!\!\!\includegraphics[width=11.5mm]{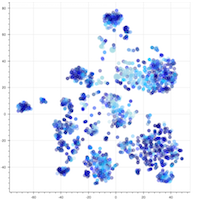}\!\!\! &
    \!\!\!\!\includegraphics[width=11.5mm]{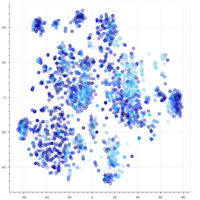} \\
    
 \!\!\includegraphics[width=8mm]{tsne/mlm.png}\!\!\! & \!\!\includegraphics[width=11.5mm]{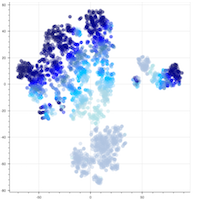}\!\!\! & \!\!\!\!\includegraphics[width=11.5mm]{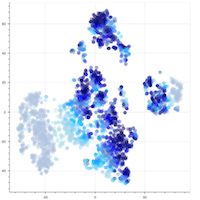}\!\!\! & \!\!\!\!\includegraphics[width=11.5mm]{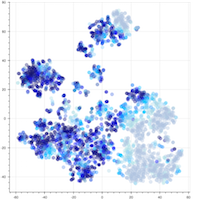}\!\!\! & \!\!\!\!\includegraphics[width=11.5mm]{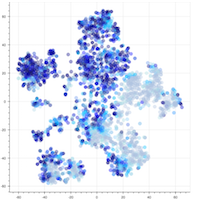}\!\!\! & 
    \!\!\!\!\includegraphics[width=11.5mm]{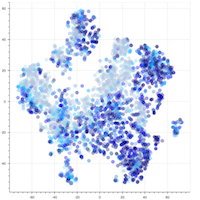}\!\!\! &
    \!\!\!\!\includegraphics[width=11.5mm]{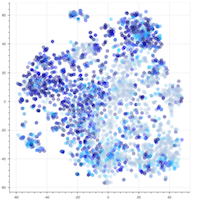} \\
\end{tabular}
\vspace{-2ex}
\caption{t-SNE of different occurrences of the token ``it'', position is in color (the larger the word index the darker its color). On the x-axis are layers.}
\vspace{-1ex}
\label{fig:tsne_it_position}
\end{figure}

We evaluate the average distance of position of the current occurrence and the top 5 closest representations. The results are provided in Figure~\ref{fig:same_position}. 
This illustrates how the information about input (in this case, position), potentially not so relevant to the output  (e.g., next word for LM), gets gradually dismissed. As expected, encoding input positions is more important for MT, so this effect is more pronounced for LM and MLM.
An illustration is in Figure~\ref{fig:tsne_it_position}. 
For MT, even on the last layer ordering by position is noticeable.

\subsection{Lexical and syntactic context}
\label{sect:evolution_past_future}

In this section, we will look at the two properties: identities of immediate neighbors of a current token and CCG supertag of a current token. On the one hand, these properties represent a model's understanding of different types of context: lexical (neighboring tokens identity) and syntactic. 
On the other, they are especially useful for our analysis since they can be split into information about `past' and `future' by taking either left or right neighbor or part of a CCG tag.

\subsubsection{The importance of neighboring tokens}
\begin{figure}[t!]
    \centering
    \begin{subfigure}[b]{0.18\textwidth}
        \includegraphics[width=\textwidth]{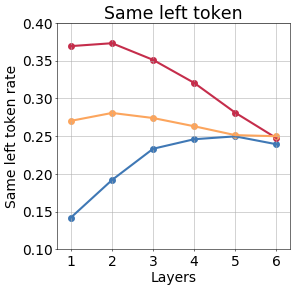}
        \caption{left}
    \end{subfigure}
    \quad
    \begin{subfigure}[b]{0.23\textwidth}
        \includegraphics[width=\textwidth]{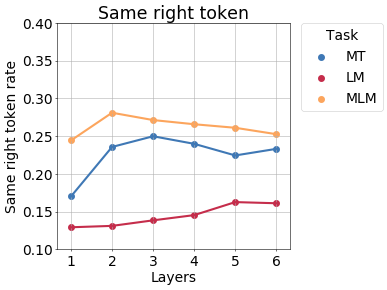}
        \caption{right}
    \end{subfigure}
    \vspace{-1ex}
    \caption{Preserving immediate neighbors
    }
    \vspace{-2ex}
    \label{fig:same_neighbor_token}
\end{figure}

Figure~\ref{fig:same_neighbor_token} supports our previous expectation that for LM the importance of a previous token decreases, while information about future token is being formed. For MLM, the importance of neighbors gets higher until the second layer and decreases after. This may reflect stages of context encoding and token reconstruction.

\begin{figure}[t!]
    \centering
    \begin{subfigure}[b]{0.15\textwidth}
        \includegraphics[width=\textwidth]{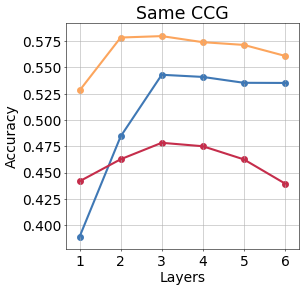}
        \caption{}
        \label{fig:same_ccg_whole}
    \end{subfigure}
    \begin{subfigure}[b]{0.15\textwidth}
        \includegraphics[width=\textwidth]{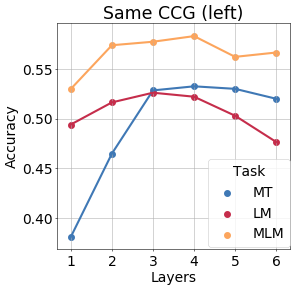}
        \caption{}
        \label{fig:same_ccg_left}
    \end{subfigure}
    \begin{subfigure}[b]{0.15\textwidth}
        \includegraphics[width=\textwidth]{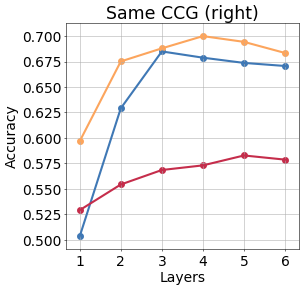}
        \caption{}
        \label{fig:same_ccg_right}
    \end{subfigure}
    \vspace{-1ex}
    \caption{Preserving CCG supertag.}
    \vspace{-2ex}
    \label{fig:same_ccg}
\end{figure}

\subsubsection{The importance of CCG tags}

Results are provided in Figure~\ref{fig:same_ccg_whole}.\footnote{To derive CCG supertags, we use \citet{yoshikawa-etal-2017-ccg} tagger, the latest version with ELMO: \url{https://github.com/masashi-y/depccg}.} As in previous experiments, importance of CCG tag for MLM degrades at higher layers. This agrees with the work by~\citet{tenney-etal-2019-bert}. The authors observe that for different tasks (e.g., part-of-speech, constituents, dependencies, semantic role labeling, coreference)  the contribution\footnote{In their experiments, representations are pooled across layers with the scalar mixing technique similar to the one used in the ELMO model~\cite{peters-etal-2018-deep}. The probing classifier is trained jointly with the mixing weights, and the learned coefficients are used to estimate the contribution of different layers to a particular task.} of a layer to a task increases up to a certain layer, but then decreases at the top layers. Our work gives insights into the underlying process defining this behavior.

For LM these results are not really informative since it does not have access to the future. We go further and measure importance of parts of a CCG tag corresponding to previous (Figure~\ref{fig:same_ccg_left}) and next (Figure~\ref{fig:same_ccg_right}) parts of a sentence. 
It can be clearly seen that LM first accumulates information about the left part of CCG, understanding the syntactic structure of the past. Then this information gets dismissed while forming information about future. 

\begin{figure}
\begin{tabular}{ccccccc} 
 \!\!\includegraphics[width=8mm]{tsne/mt.png}\!\!\! & \!\!\includegraphics[width=11.5mm]{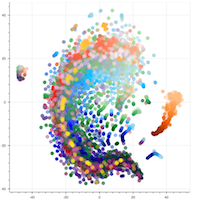}\!\!\! & \!\!\!\!\includegraphics[width=11.5mm]{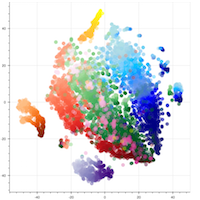}\!\!\! & \!\!\!\!\includegraphics[width=11.5mm]{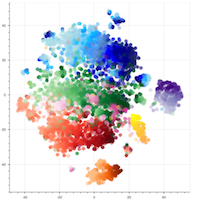}\!\!\! & \!\!\!\!\includegraphics[width=11.5mm]{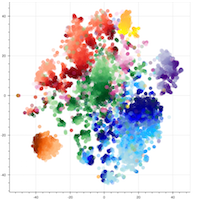}\!\!\! & 
    \!\!\!\!\includegraphics[width=11.5mm]{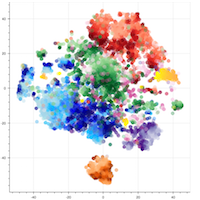}\!\!\! &
    \!\!\!\!\includegraphics[width=11.5mm]{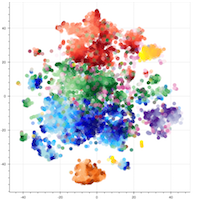} \\
    
 \!\!\includegraphics[width=8mm]{tsne/lm.png}\!\!\! & \!\!\includegraphics[width=11.5mm]{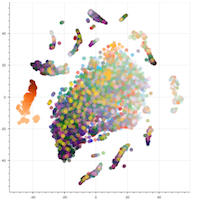}\!\!\! & \!\!\!\!\includegraphics[width=11.5mm]{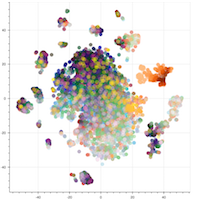}\!\!\! & \!\!\!\!\includegraphics[width=11.5mm]{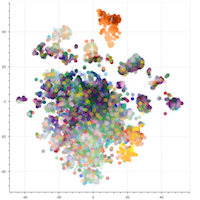}\!\!\! & \!\!\!\!\includegraphics[width=11.5mm]{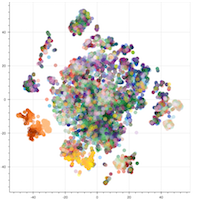}\!\!\! & 
    \!\!\!\!\includegraphics[width=11.5mm]{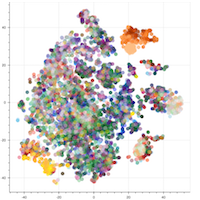}\!\!\! &
    \!\!\!\!\includegraphics[width=11.5mm]{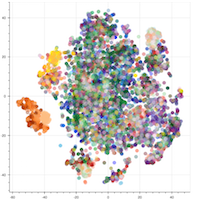} \\
    
 \!\!\includegraphics[width=8mm]{tsne/mlm.png}\!\!\! & \!\!\includegraphics[width=11.5mm]{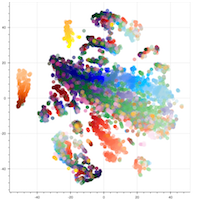}\!\!\! & \!\!\!\!\includegraphics[width=11.5mm]{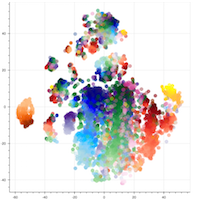}\!\!\! & \!\!\!\!\includegraphics[width=11.5mm]{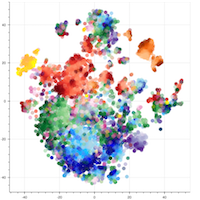}\!\!\! & \!\!\!\!\includegraphics[width=11.5mm]{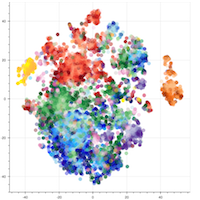}\!\!\! & 
    \!\!\!\!\includegraphics[width=11.5mm]{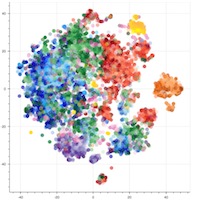}\!\!\! &
    \!\!\!\!\includegraphics[width=11.5mm]{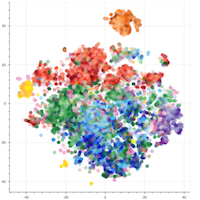} \\
\end{tabular}
\vspace{-1ex}
\caption{t-SNE of different occurrences of the token ``is'', CCG tag is in color (intensity of a color is a token position). On the x-axis are layers.}
\vspace{-2ex}
\label{fig:tsne_is_ccg}
\end{figure}

Figure~\ref{fig:tsne_is_ccg} shows how representations of different occurrences of the token ``is'' get reordered in the space according to CCG tags (colors correspond to tags).

\section{Additional related work}

Previous work analyzed representations of MT and/or LM models by 
using probing tasks.
Different levels of linguistic analysis have been considered including morphology~\cite{P17-1080,dalvi-morph-decoder,bisazza-tump:2018:EMNLP}, syntax~\cite{D16-1159,tenney2018you} and semantics~\cite{hill-et-al,belinkov-I17-1001,raganato-tiedemann:2018:BlackboxNLP,tenney2018you}. 
Our work complements this line of research by analyzing how word representations evolve between layers 
and gives insights into how models trained on different tasks 
come to represent different information.

% CCA
Canonical correlation analysis has been previously used to investigate learning dynamics of CNNs and RNNs, to measure the intrinsic dimensionality of layers in CNNs and compare representations of networks which memorize and generalize~\cite{svcca-2017,pwcca-2018}.
\citet{bau2019neurons-in-mt} used SVCCA as one of the methods used for identifying important individual neurons in NMT models. \citet{saphra-lopez-2019-understanding} used SVCCA to investigate how representations of linguistic structure are learned over time in LMs.

\section{Conclusions}
In this work, we analyze how the learning objective determines the information flow in the model. 
We propose to view the evolution of a token representation between layers from the compression/prediction trade-off perspective.
We conduct a series of experiments 
    supporting this view and 
    propose a possible explanation for superior performance of MLM over LM for pretraining.
    We relate our findings to observations previously made in the context of probing tasks.

\section*{Acknowledgments}
We would like to thank the anonymous reviewers for their comments.
The authors also thank Artem Babenko, David Talbot and Yandex Machine Translation team for helpful discussions and inspiration. Ivan Titov acknowledges support of the European Research Council (ERC StG BroadSem 678254) and the Dutch National Science Foundation (NWO VIDI 639.022.518). 
Rico Sennrich acknowledges support from the Swiss National Science Foundation (PP00P1\_176727).

\nocite{sennrich-bpe}
\nocite{manning-EtAl:2014:P14-5}

\bibliography{emnlp-ijcnlp-2019}
\bibliographystyle{acl_natbib}

\newpage
\appendix

\section{Data and Setting}

\subsection{Model architecture}
For machine translation models, we follow the setup of the Transformer base model~\cite{attention-is-all-you-need}. More precisely, the number of layers in the encoder and in the decoder is $N=6$. We use $h = 8$ parallel attention layers, i.e. heads. The dimensionality of input and output is $d_{model} = 512$, and the inner-layer of the feed-forward networks has the dimensionality $d_{ff}=2048$.
For language models, we use only the encoder of the model (with the same hyper-parameters).

\subsection{Training}
Sentences were encoded using byte-pair encoding~\cite{sennrich-bpe}, with source and target vocabularies of about 32000 tokens. Source vocabulary is the same for all tasks.
Minibatch size is set to approximately 15000 source tokens. Training examples were batched together by approximate sequence length. Each training batch contained a set of translation pairs, approximately 15000 source tokens each.
The optimizer and learning rate schedule we use are the same as in~\citet{attention-is-all-you-need}. Since using a large number of training steps was reported to be important for the MLM objective,
 we follow~\citet{bert} and train MLM for 1 million training steps and other models till convergence.

\section{Fine-grained analysis of change and influence: varying PoS}
\label{sect:appendix:pos}

\begin{figure}[t!]
    \centering
    \begin{subfigure}[b]{0.21\textwidth}
        \includegraphics[width=\textwidth]{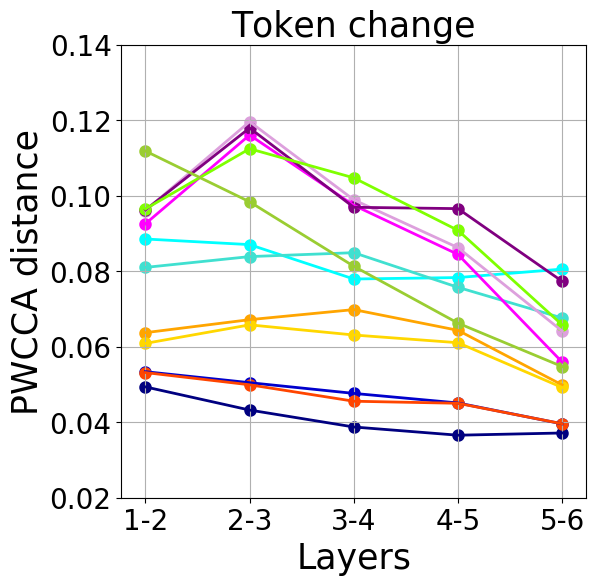}
        \caption{MT}
        \label{fig:change_by_postag_mt}
    \end{subfigure}
    \begin{subfigure}[b]{0.21\textwidth}
        \includegraphics[width=\textwidth]{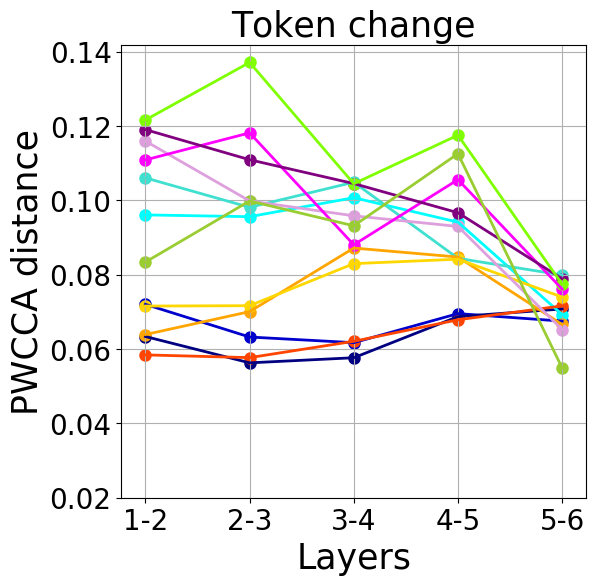}
        \caption{LM}
        \label{fig:change_by_postag_lm}
    \end{subfigure}
    \begin{subfigure}[b]{0.21\textwidth}
        \includegraphics[width=\textwidth]{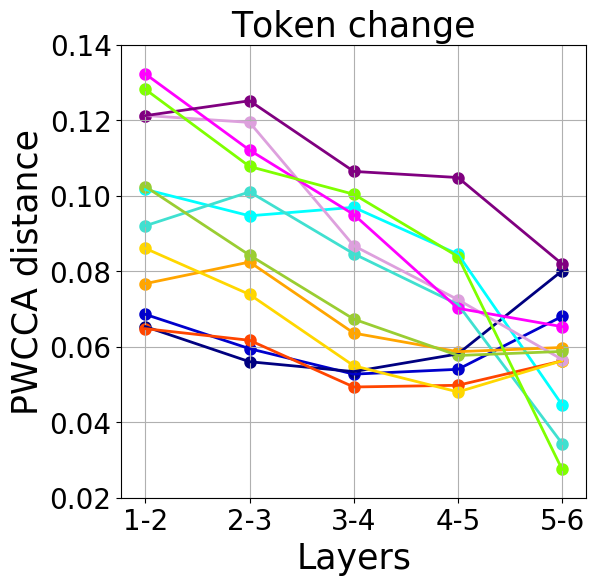}
        \caption{MLM}
        \label{fig:change_by_postag_mlm}
    \end{subfigure}
    \begin{subfigure}[b]{0.21\textwidth}
        \includegraphics[width=\textwidth]{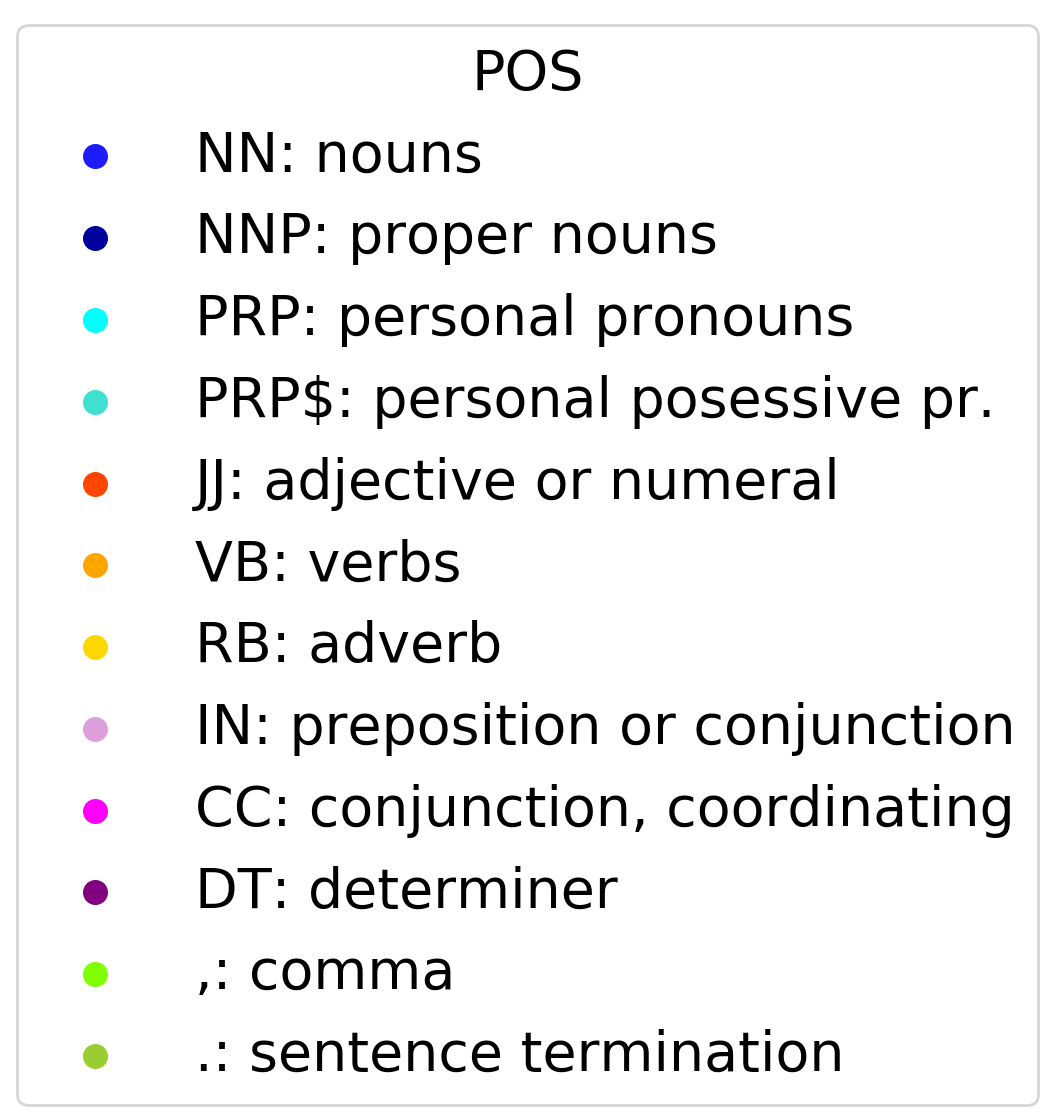}
    \end{subfigure}
    \vspace{-1ex}
    \caption{Token change vs its part of speech.}\label{fig:change_by_postag}
    \vspace{-1ex}
\end{figure}

\begin{figure}[t!]
    \centering
    \begin{subfigure}[b]{0.21\textwidth}
        \includegraphics[width=\textwidth]{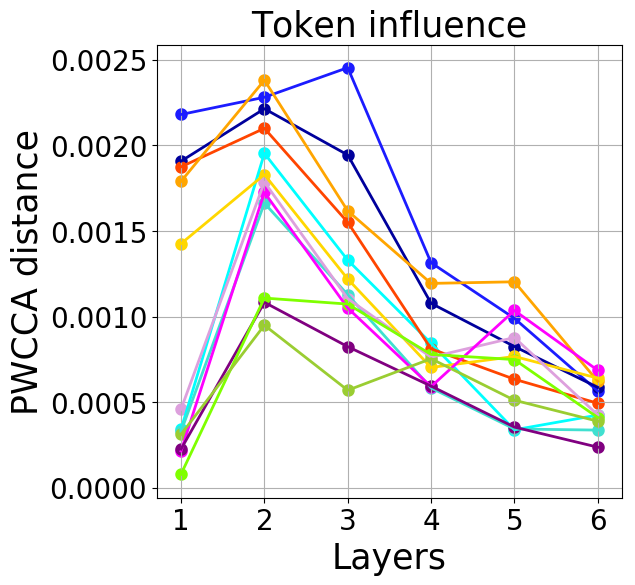}
        \caption{MT}
        \label{fig:influence_by_postag_mt}
    \end{subfigure}
    \begin{subfigure}[b]{0.21\textwidth}
        \includegraphics[width=\textwidth]{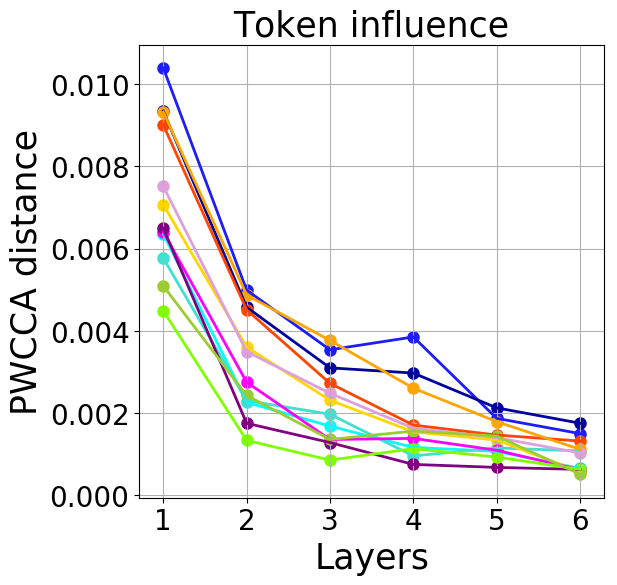}
        \caption{LM}
        \label{fig:influence_by_postag_lm}
    \end{subfigure}
    \begin{subfigure}[b]{0.21\textwidth}
        \includegraphics[width=\textwidth]{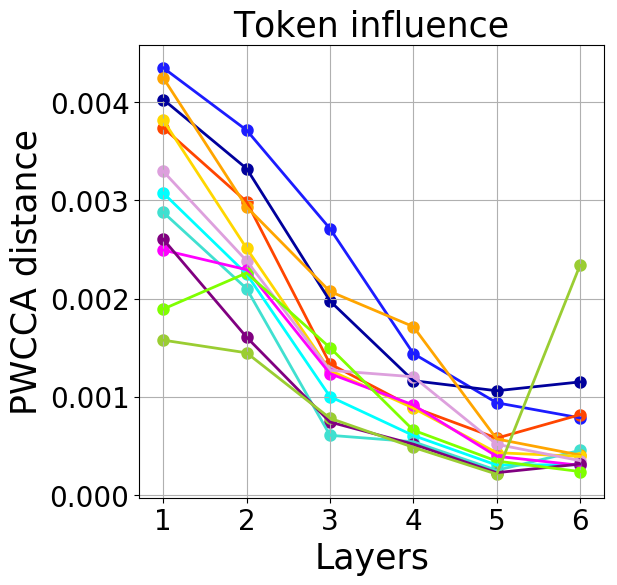}
        \caption{MLM}
        \label{fig:influence_by_postag_mlm}
    \end{subfigure}
    \begin{subfigure}[b]{0.21\textwidth}
        \includegraphics[width=\textwidth]{pict/postag_wmt_en_de_bbox_no_eos.png}
    \end{subfigure}
    \vspace{-1ex}
    \caption{Token influence vs its part of speech.}\label{fig:influence_by_postag}
    \vspace{-2ex}
\end{figure}

Figure~\ref{fig:change_by_postag} shows  the amount of change for different parts of speech, Figure~\ref{fig:influence_by_postag} -- the amount of influence for different parts of speech\footnote{We use the part-of-speech tagger from Stanford CoreNLP~\cite{manning-EtAl:2014:P14-5}.}. Generally, the patterns are similar to the ones for frequency groups: parts of speech with frequent tokens (preposition, conjunction, etc.) change more and influence less.

\section{What does a layer represent?}
\subsection{Preserving token identity: experimental setup}

In this section, we want to check to what extent a model confuses representations of different tokens. For each of the selected tokens described above (``main'' tokens), we pick tokens which potentially could be confused with the token under consideration (``contrastive'' tokens). In our setting, contrastive tokens are the top 10 closest tokens in the embedding space (separately for each model).  For example, tokens ``is'', ``are'', ``was'', ``were'' are close in the embedding space of each of the models. Then for each main token we gather representations of 9000 different occurrences of its contrastive tokens, and add them to the 1000 states of the main token. All in all, we get 200 groups of representations; each group contains representations of 1000 occurrences of the main token and 9000 of the contrastive ones (at each layer). 
For representations of the main token in a group, we measure the average percentage of representations of the same token for top closest among 10000 representations.

\end{document}